\documentclass{elsarticle}

\usepackage{amsmath}
\usepackage{amssymb}
\usepackage{mathtools}
\usepackage{multirow}
\usepackage{stmaryrd}
\usepackage{multirow}
\usepackage{etoolbox}
\usepackage[hyphens]{url}
\usepackage[breaklinks]{hyperref}
\usepackage{graphicx}
\usepackage{booktabs} 
\usepackage{rotating}
\usepackage[bottom]{footmisc}

\usepackage{float}
\usepackage{caption}
\usepackage{subcaption}
\usepackage{placeins}
\usepackage[T1]{fontenc}

\def\ptFiguresDirectory#1{./figures/#1}
\def\FWER#1{FWER}
\def\fdr{\mathrm{FDR}}
\def\fnr{\mathrm{FNR}}

\def\mcc{\mathrm{MCC}}

\def\neighCoeff{\zeta}
\def\dotProd#1#2{\left\langle {#1};{#2}\right\rangle}
\def\vnorm#1{\left\lVert {#1} \right\rVert}
\DeclareMathOperator{\sign}{sign}
\def\setBracks#1{\left\{{#1}\right\}}

\def\vec#1{\mathbf{#1}}
\def\rndVar#1{\mathbf{#1}}

\def\posPotential{\beta}
\def\newPotential{\epsilon}
\def\distas{\sim}

\begin{document}

\begin{frontmatter}


\title{Probability-driven scoring functions in combining linear classifiers}
\author{Pawel Trajdos\corref{cor1}}
\author{Robert Burduk}


\cortext[cor1]{pawel.trajdos@pwr.edu.pl}
\address{Department of Systems and Computer Networks, Wroclaw
University of Science and Technology, \\ Wybrzeze Wyspianskiego 27, 50-370
Wroclaw, Poland }

\begin{abstract}
Although linear classifiers are one of the oldest methods in machine learning, they are still very popular in the machine learning community. This is due to their low computational complexity and robustness to overfitting. Consequently, linear classifiers are often used as base classifiers of multiple ensemble classification systems. This research is aimed at building a new fusion method dedicated to the ensemble of linear classifiers.  The fusion scheme uses both measurement space and geometrical space. Namely, we proposed a probability-driven scoring function which shape depends on the orientation of the decision hyperplanes generated by the base classifiers. The proposed fusion method is compared with the reference method using multiple benchmark datasets taken from the KEEL repository. The comparison is done using multiple quality criteria. The statistical analysis of the obtained results is also performed.  The experimental study shows that, under certain conditions, some improvement may be obtained.
\end{abstract}

 \begin{keyword}
 Linear Classifier \sep Potential Function \sep Ensemble of Classifiers \sep Score Function
 \end{keyword}

\end{frontmatter}

\section{Introduction}\label{sec:Instroduction}

The concept of a linear classifier is one of the oldest machine learning methods. They are dated back to the 1950s~\cite{Rosenblatt1958}. For decades many algorithms for building linear classifiers have been developed~\cite{McLachlan1992,Devroye1996,gurney1997an,Kuncheva1998,Cortes1995,Zhu2018}. Today they are still used by the machine learning society~\cite{kim2017object}. This is due to the relatively low computational complexity of training and predicting phases~\cite{plumpton2012naive}. What is more, after a suitable data preprocessing, linear classifiers may offer classification quality comparable to nonlinear ones~\cite{Yuan2012}. Additionally, for some linear classifiers, it is possible to obtain nonlinear decision boundaries using the kernel trick~\cite{Zhu2018}. Nonlinear decision boundaries may also be obtained using ensembles of classifiers~\cite{kim2017object,Taud2017}. 

Ensemble classifiers are complex systems built with multiple classifiers called base classifiers. That is why they are also called multiclassifiers~\cite{pt:Wozniak2014}. The classifiers constituting the ensemble are trained together, and then their outputs are combined to form the response of the ensemble~\cite{Dong2019}. Using multiple classifiers instead of one has many advantages. As it was mentioned above, combining multiple classifiers allows building a more complex decision boundary~\cite{kim2017object}. What is more, it also improves the result stability (reduces variance) and robustness to outliers~\cite{Sagi2018}. Consequently, in general, they allow obtaining better classification quality compared to a single model~\cite{Krawczyk2017}. They proved to be efficient tools for dealing with a wide range of practical classification tasks~\cite{Zhang2021,Yang2021,Chandra2021,Mehmood2021}. This is the reason why building an ensemble of classifiers is still a widely explored topic in machine learning~\cite{pt:Wozniak2014,Krawczyk2017, Sagi2018, Dong2019}.

The operation of the ensemble classifier is usually divided into two stages: generation and integration (fussion)~\cite{mohandes2018classifiers}. During the generation phase, a set of accurate and diverse classifiers is constructed. By exact, we mean the classifier which accuracy is better than the random guessing. Simply speaking, the diversity guarantees that different base classifiers make invalid predictions for different samples. Diversity is more important because we cannot benefit from combining a set of identical classifiers~\cite{pt:Dietterich2000}.  There are two well-known ways of creating a diversified group of classifiers. One involves building an ensemble of classifiers based on various learning paradigms (heterogeneous ensemble)~\cite{GhaderiZefrehi2020}.  The other is to build a homogeneous ensemble utilizing classifiers that follow the same learning paradigm, but the models are trained using different training data~\cite{Wang2021}. The most commonly employed methods of creating homogeneous ensembles are bagging~\cite{pt:breiman1996}, boosting~\cite{Freund1996}, random subspaces~\cite{pt:Ho1998}, and random projections~\cite{Vrahatis_2020}. 

In the integration stage, the final decision of the ensemble is obtained. In this stage, a certain subset of previously created base classifiers can be selected~\cite{Cruz2018}. Generally, the fusion strategies may be divided into a few categories~\cite{pt:Rokach2010,pt:Wozniak2014} 
  \begin{itemize}
   \item Trainable~\cite{pt:Kuncheva2014} and untrainable~\cite{RB:KittlerA03}. The trainable ones need the combiner to be trained using a separate set of training data, whereas the untrainable ones do not;
   \item Static~\cite{pt:Kuncheva2014, RB:KittlerA03} and dynamic~\cite{pt:Valdovinos2009}.  Static combiners use the same mix of base classifiers for each example. On the other hand, the dynamic combination procedure depends on the sample being classified for the dynamic ones. 
\end{itemize}
The classifier fusion can be done using various output spaces~\cite{mohandes2018classifiers,Pujol2009}.
\begin{itemize}
 \item Abstract space (class space) combiners using only information about the class assigned to the sample~\cite{kuncheva2014combining}.
 \item Rank space. In this case, the base classifiers produce a class ranking. The position within the ranking is expressed using an integer number~\cite{PrzybyaKasperek2017}. 
 \item The measurement space. In the measurement space, the classifier expresses class-specific support using a real number. The higher the number is, the greater is belief that the sample belongs to the given class. The class support values are usually normalized within the $[0;1]$ interval and sum up to one~\cite{kuncheva2014combining}.
 \item Geometric space. The geometric combiners use properties of the geometrical space in which the decision boundary is placed~\cite{Pujol2009}. For example, the method proposed in~\cite{Pujol2009} uses characteristic boundary points and weight estimation to provide a piecewise linear classifier. On the other hand, the methods proposed in~\cite{LiYujian2011,Leng2019} constructs a multiconlitron that separates the classes.  
\end{itemize}

In this paper, we extended the probability-based scoring function proposed in~\cite{Trajdos2020a}. The scoring function combines measurement-level fusion with geometric space fusion. That is, it uses geometric properties of the space combined with the probabilistic framework. We have proposed three extended scoring functions. The scoring functions harness additional probabilistic information about the data distribution in the input (geometric) space. Namely, they utilize information about point distribution along the decision plane. The function proposed in~\cite{Trajdos2020a} uses only information about the point spread along the normal vector of the decision plane. 

The main objectives of this work can be summarized as follows:
\begin{itemize}
 \item A proposal of new scoring functions that better utilize the probabilistic information available. 
 
 \item Harnessing the proposed scoring functions in the task of building a homogeneous ensemble of classifiers. 
 
 \item An experimental setup to compare the proposed scoring functions with the reference methods. The comparison is done in terms of the selected quality criteria. During the experiments, different base classifiers are used. 
 
\end{itemize}

The outline of the paper is as follows: In the next section (Section~\ref{sec:RelWork}), related works are outlined. The proposed methods are presented in Section~\ref{sec:Proopsed}. In Section \ref{sec:ExpEval}, the research questions are formulated, and the experimental setup is described. The experimental results are presented and discussed in Section~\ref{sec:ResAndDisc}. Finally, the paper is concluded in section~\ref{sec:Conclusions}.

\section{Related Work}\label{sec:RelWork}

This section describes the previous work related to the problem of building ensembles of linear classifiers. We begin with the definition of a linear classifier and then switch to the topic of building ensembles of such classifiers.

\subsection{Linear Classifier}\label{sec:RelWork:LinearClassifier}

Let us begin with the definition of a linear classifier. The linear classifier assigns points taken from the feature space $\mathbb{X}$, which in this work is assumed to be a $d-\mathrm{dimensional}$ Euclidean space $\mathbb{X}=\mathbb{R}^d$, to two possible classes $\mathbb{M}=\left\{-1;1\right\}$~\cite{duda2012pattern}. To separate the two classes of feature space points, the classifier utilizes a hyperplane $\pi$ defined by the following equation:
\begin{align}
 \pi: \dotProd{\vec{n}}{\vec{x}} + b &=0,
\end{align}
where $\vec{n}$ is a unit normal vector of the decision hyperplane, $b$ is the distance from the hyperplane to the origin and $\dotProd{\cdot}{\cdot}$ is a dot product defined as follows~\cite{Kostrikin2005}:
\begin{align}
 \dotProd{\vec{a}}{\vec{b}} = \sum_{i=1}^{d}a_{i}b_{i}\;\forall \vec{a},\vec{b} \in \mathbb{X}.
\end{align}
The norm of the vector $\vec{x}$ is defined using the dot product:
\begin{align}\label{eq:dotPDist}
 \vnorm{\vec{x}} &= \sqrt{\dotProd{\vec{x}}{\vec{x}}}.
\end{align}  

For each instance $\vec{x}$, the linear classifier $\psi$ produces the discriminant function~\cite{kuncheva2014combining}:
\begin{align}
 \omega(\vec{x}) &= \dotProd{\vec{n}}{\vec{x}} + b,
\end{align}
which absolute value equals the perpendicular distance from the decision hyperplane $\pi$ to the point $\vec{x}$. The sign of the value returned by the discriminant function depends on the site of the plane where the instance $\vec{x}$ lies. The decision of the linear classifier is thus determined by checking the sign of the discriminant function:
\begin{align}\label{eq:LinClass}
 \psi(\vec{x}) &= \sign \left( \omega(\vec{x}) \right).
\end{align}

 During the training phase of the linear classifier, the proper decision plane is found using the training set $\mathcal{T}$. which consists of $|\mathcal{T}|$ (where $|\cdot|$ is the cardinality of a set) pairs of feature space vectors $\vec{x}$ and their corresponding class labels ${m}$:
\begin{equation}\label{eq:trainSet} 
\mathcal{T}=\left\{({\vec{x}}^{(1)},m^{(1)}), ({\vec{x}}^{(2)},m^{(2)}), \ldots ,({\vec{x}}^{(|\mathcal{T}|)},m^{(|\mathcal{T}|)})\right\},
\end{equation}
where ${\vec{x}}^{(k)} \in \mathbb{X}$ and $m^{(k)} \in \mathbb{M}$. The literature contains various procedures for obtaining the decision plane~\cite{Yuan2012}. Among the others, we may mention such algorithms as: FLDA~\cite{McLachlan1992}, Logistic Regression~\cite{Devroye1996}, Perceptron~\cite{gurney1997an}, Nearest Centroid Classifier~\cite{Kuncheva1998}, and SVM~\cite{Cortes1995}. The procedures used in the experiments are listed in Section~\ref{sec:ExpEval:setup}.

Despite their simplicity, linear classifiers are often used to solve practical classification tasks~\cite{Yuan2012}. First of all, they are useful due to their low computational complexity. Additionally, due to their simplicity, they are also less overfitting prone~\cite{plumpton2012naive}. What is more, they can obtain a classification quality comparable to nonlinear classifiers when the dimensionality of the input space is high~\cite{Yuan2012}. However, for some classification problems with nonlinear classification boundaries they are insufficient~\cite{Zhu2018}. One solution may be to tailor linear classifiers to find non-linear decision boundary. This may be done by applying the kernel trick~\cite{Yuan2012,Zhu2018}. The other way is to build a structure that consists of multiple linear classifiers that are trained together. An example of such a technique is to build a multilayer neural network~\cite{Taud2017}, a deep neural network in particular~\cite{Begum2019}. Another way is to use the multi-classifier approach and build an ensemble of linear classifiers~\cite{kim2017object}.

\subsection{An Ensemble of Linear Classifiers}\label{sec:RelWork:Ensembles}

Generally speaking, an ensemble of classifiers (a multiclassifier) is a set of classifiers that work together to deliver more robust results~\cite{kuncheva2014combining}. Throughout this paper, the ensemble of classifiers is denoted by:
\begin{align}
  {\Psi} &=\setBracks{\psi^{(1)},\psi^{(2)},\cdots,\psi^{(N)}}
\end{align}

In this paper, we are focused on the ensemble combination methods dedicated to linear classifiers. The proposed weighting methods are trainable and dynamic ones that combine the base classifiers in the geometric space. That is, the weights depend on the orientation of the decision plane. 

In the literature, we may find multiple methods of combining linear classifiers. Now, we list these methods starting from the simplest one. The most straightforward way to combine the results of several classifiers is to use model averaging~\cite{Skurichina1998}. The model averaging approach is to simply calculate the mean value of the classifier-specific discriminant functions:
\begin{align}
 \label{eq:softVoting} \omega(\vec{x}) &= \frac{1}{N}\sum_{i=1}^{N}\omega^{(i)}(\vec{x}),
\end{align}
where $\omega^{(i)}(\vec{x})$ is the value of the discriminant function provided by the classifier $\psi^{(i)}$ for the point $\vec{x}$. As we said before, the value taken by the discriminant function of the linear classifier is proportional to the distance from the given point $\vec{x}$ to the decision plane. In general, this distance is unbounded, which poses a major disadvantage of the model averaging approach. That is, when one of the base classifiers has produced a misplaced decision boundary, the high value of the discriminant function coming from this boundary may significantly change the response of the entire ensemble. 

This issue may be easily addressed by ignoring the exact value of the discriminant function and taking only the sign of the value. This approach is called majority voting, and the response of the ensemble is given by following formula~\cite{alpaydin2020introduction}:
\begin{align}
 \label{eq:crispVoting}\omega(\vec{x}) &= \sum_{i=1}^{N}\sign\left[\omega^{(i)}(\vec{x})\right],
\end{align}
Although this approach is robust to misplaced decision boundaries, it loses the information related to the exact value of the discriminant function.

The aforementioned disadvantage can be partly eliminated by the application of a type of a sigmoid transformation~\cite{kuncheva2014combining}. The sigmoid function, also called S-shaped function, is an increasing one that has finite upper and lower bounds. An example of such a function is the softmax function:
\begin{align}
 \label{eq:softMax} \widetilde{\omega}^{(i)}(\vec{x}) &= \left( 1+ \exp \left[-\omega^{(i)}(\vec{x})\right] \right)^{-1}.
\end{align}
Applying this kind of transformation ensures that distance-specific information is not lost, and it also reduces the impact of misplaced hyperplanes. Employing a simple sigmoid function is a simplified version of the probability calibration task. In this task, we want to provide an estimation of the class posterior probability distribution~\cite{Kull2017}.  This distribution may be obtained using various techniques such as Plat scaling (logistic calibration)~\cite{platt1999probabilistic,Bken2021}, sigmoid fitting~\cite{Zadrozny2002}, or beta scaling~\cite{Kull2017}
To produce the final outcome of the ensemble, the transformed values are averaged: 
\begin{align}
 \label{eq:softVoting2} \omega(\vec{x}) &=\frac{1}{N}\sum_{i=1}^{N}  \widetilde{\omega}^{(i)}(\vec{x}).
\end{align}

As it was said before, one of the problems with combining the linear classifiers is that the discriminative function grows monotonically with the distance to the decision plane. This poses no problem when a single classifier is queried. However, it may cause a situation where the absolute value of the discriminant function is high, but the training set of a classifier contains no instances so far from the decision boundary. In other words, the classifier returns a high value of the discriminant function outside its region of competence, which may distort the final prediction of the ensemble. The application of a sigmoid function or more generally a sort of posterior probability scaling mitigates the problem, but does not resolve it. The reason is that at a great distance from the decision boundary, the calibrated discriminant function approaches its upper (lower) limit. The values close to the limits still express relatively high class-specific support outside the competence region of the base classifier. Our previous research has shown that reducing the value of the discriminant function outside the competence region of the classifier may significantly improve the classification quality achieved by the ensemble~\cite{Trajdos2019,Trajdos2020a}. Our first attempt was to provide a simple non-monotonic parametric function~\cite{Trajdos2019}:
\begin{align}\label{eq:Potential2}
 g\big(\omega(\vec{x})\big) &= \omega(\vec{x})\exp\left[-\neighCoeff \big(\omega(\vec{x})\big)^{2} +0.5\right]\sqrt{2\neighCoeff},
\end{align}
where $\neighCoeff$ is a coefficient that controls the position and steepness of peaks. Unfortunately, we have not proposed a closed-form formula for finding the good value of this coefficient. Consequently, the proper value of this coefficient must be found using cross-validation. The translation constant $0.5$ and the scaling factor $\sqrt{2\neighCoeff}$ assure that the maximum and positive and negative peaks of the discriminant functions are $1$ and $-1$ respectively. The final value of the discriminant function of the ensemble is calculated by averaging the transformed values given by the base classifiers:
\begin{align}\label{eq:transResponse}
 \omega(\vec{x}) & = \frac{1}{N}\sum_{i=1}^{N}g\left(\omega^{(i)}(\vec{x})\right).
\end{align}
The conducted experimental evaluation showed that applying this kind of non-monotonic transformation causes a gain in the classification quality obtained by heterogeneous ensembles.  Unfortunately, the practical applications of this method are limited since it is very sensitive to imbalanced class distribution. What is more the $\neighCoeff$ coefficient has to be tuned for each dataset separately. To eliminate these drawbacks, we proposed an approach that models the data spread along the plane vector using kernel probability estimators~\cite{Trajdos2020a}. The conducted experimental evaluation showed that the previously proposed method offers some improvement over the formerly proposed and reference methods.

\begin{figure}[tb]
 \centering
  \includegraphics[width = 0.8\textwidth, height=.3\textheight, keepaspectratio]{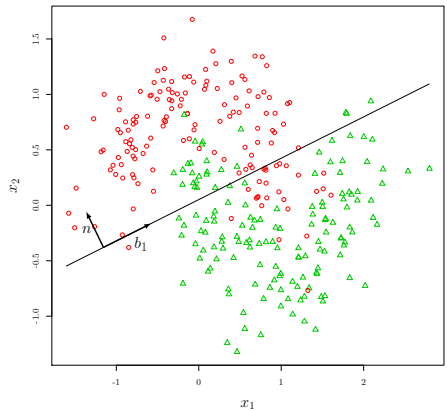}
  \caption{The linear decision boundary for binary, two-dimensional data. Objects belonging to the first class have been marked using red circles, whereas points belonging to the other class have been marked using green triangles. The plot also shows the normal vector of the decision plane $n$ and a base vector of the plane $b_1$.  \label{fig:LinClassSim}}
\end{figure}

The discriminant function created by a linear classifier uses only the information about the distance between an object and the decision hyperplane. However, the information about the data distribution along the basis of the decision hyperplane may also be useful when determining the competence region of the base classifier.  The basis of the decision hyperplane is a set of linearly independent vectors that span the plane~\cite{Kostrikin2005}:
\begin{align}
 B &= \setBracks{\vec{b}_1,\vec{b}_2,\cdots, \vec{b}_{d-1}}.
\end{align}
The plane base for the two-dimensional classification problem is shown in \figurename~\ref{fig:LinClassSim}.
 
With that in mind, we proposed a method that incorporates this information into the ensemble classifier.

\section{Proposed Improvements of the Probability-based Potential function}\label{sec:Proopsed}

Let us begin with a more detailed description of the method proposed in~\cite{Trajdos2021a}. This description is needed since the methods proposed in this paper are extensions of the above-mentioned methods. 

\subsection{Potential Functions}\label{sec:Proopsed:PotFun}

The potential function defined in~\cite{Trajdos2021a} is defined using a probabilistic framework. It means that $\vec{x}$ and $m$ are realizations of random variables $\rndVar{X}$ and $\rndVar{M}$, respectively. The joint distribution $P(\rndVar{X}, \rndVar{M})$ is also known. Then, the value of the discriminant function $\omega(\vec{x})$ is also a realisation of a random variable defined as follows:
\begin{align}
 \rndVar{W} &= \dotProd{\vec{n}}{\rndVar{X}} + b.
\end{align} 
This one-dimensional distribution describes the data spread along the line defined by the normal vector $n$ of the decision plane. This random variable is also jointly distributed with $\rndVar{M}$: $P(\rndVar{W}, \rndVar{M})$.  We denote its probability density function of this random variable by $w(\omega)$.

Under these assumptions, we may define the conditional probability of class $m=1$ given $\omega(\vec{x})$:
\begin{align}
 P\Big(\rndVar{M}=1|\omega(\vec{x})\Big)  &= \frac{w\Big(\omega(\vec{x})|\rndVar{M}=1\Big)P(\rndVar{M}=1)}{ \sum_{m\in\mathbb{M}}w\Big(\omega(\vec{x})|\rndVar{M}=m\Big)P(\rndVar{M}=m)}.
\end{align}
The potential function is then defined to be proportional to the probability: 
\begin{align}\label{eq:pot1}
 \posPotential\Big(\omega(\vec{x})\Big) &= \frac{ \exp\Big[w\big(\omega(\vec{x})|\rndVar{M}=1\big)P(\rndVar{M}=1)\Big]}{ \sum_{{{m\in\mathbb{M}}}} \exp\Big[w\big(\omega(\vec{x})|\rndVar{M}=m\big)P(\rndVar{M}=m)\Big]} -0.5.
\end{align}
Using the softmax transformation allows avoiding numerical problems in areas with low point density. Subtracting $0.5$ from the expression puts the result into $[-0.5;0.5]$ interval. 

In this work, we employed probability density estimations that describe the point distribution along the vectors of the plane basis $B$. To do so, we define a new multidimensional random variable which elements are defined as the projection coefficients of the random variable $\rndVar{X}$ onto the base vectors of the decision plane:
\begin{align}\label{eq:PlaneBaseDistr}
 \rndVar{Y}_i &= \frac{\dotProd{\rndVar{X}}{\vec{b}_i}}{\vnorm{\vec{b}_i}}.
\end{align}
Consequently, the random variable describes the distribution of points projected onto the basis of the decision plane. The probability density function of this random variable is denoted by $y(\vec{x})$. 

\begin{figure}
\centering
\includegraphics[width=.7\textwidth, height=.3\textheight, keepaspectratio]{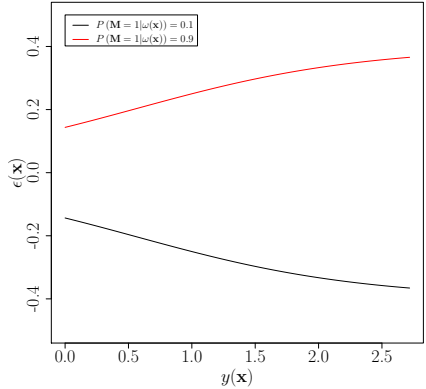}
 \caption{The plot of potential functions for different values of $P\Big(\rndVar{M}=1|\omega(\vec{x})\Big)$. The $y(\vec{\mu})$ value is set to 1.0. }
 \label{fig:potential1}
\end{figure}

Our first approach is to define the modified potential function using $y(x)$ solely. This potential function uses the information about the point distribution regardless the class assigned to each of the points. We assumed that $\rndVar{Y}$ is normally distributed with the expected value $\vec{\mu}$ and the covariance matrix $\Sigma$: $\rndVar{Y} \distas \mathcal{N}(\mu,\Sigma)$. This assumption is made because the normal distribution is a unimodal one. Consequently, the max value of the probability density function can be easily determined. The potential function is then defined as: 
\begin{align}\label{eq:newPotential1}
 \newPotential(\vec{x}) &= \frac{1}{z(\vec{x})}P\Big(\rndVar{M}=1|\omega(\vec{x})\Big)^{\frac{\exp{\left(y(\vec{x})\right)}}{\exp{\left(y(\vec{x})\right) + \exp{\left(y(\mu)\right)}}}} -0.5,
\end{align}
where $z(\vec{x})$ is a normalization factor that guarantees these potentials comming from $P\Big(\rndVar{M}=1|\omega(\vec{x})\Big)$ and $P\Big(\rndVar{M}=-1|\omega(\vec{x})\Big)$ sum up to zero. Exponent in the equation~\eqref{eq:newPotential1} is a softmax between the highest pdf value of the distribution $y(\vec{\mu})$ and $y(\vec{x})$. For low values of $y(\vec{x})$ the value of the potential function tends to 0. Consequently, in the areas where the concentration of samples is low, the discriminant function of the base classifier is close to zero. In other words, in those areas, the classifier cannot definitely say which class to choose. An example plot of the potential function is shown in~\figurename~\ref{fig:potential1} . 

Another strategy is to use conditional probabilities $y(\vec{x} | \rndVar{M}=m)$ and use them to calculate $P\Big(\rndVar{M}=1|\omega(\vec{x})\Big)$. To employ these probabilities, we made a naive assumption that $y(\vec{x} | \rndVar{M}=m)$ and $w\Big(\omega(\vec{x})|\rndVar{M}=m\Big)$ are conditionally independent given $\rndVar{M}=m$. The same assumption is done in the Naive Bayes classifier~\cite{Hand2001}. Taking this into account, the conditional class probability is calculated using the following formula:
\begin{align}\label{eq:ClassCondProbEstim2}
 P\Big(\rndVar{M}=1|\omega(\vec{x})\Big)  &= \frac{w\Big(\omega(\vec{x})|\rndVar{M}=1\Big) y\Big(\vec{x}|\rndVar{M}=1 \Big) P(\rndVar{M}=1)}{ \sum_{m\in\mathbb{M}}w\Big(\omega(\vec{x})|\rndVar{M}=m\Big)y\Big(\vec{x}|\rndVar{M}=m \Big)P(\rndVar{M}=m)}.
\end{align}
Consequently, the potential is calculated as follows:
\begin{align}\label{eq:potN2}
 \newPotential_{2}(x) &= \frac{ \exp\Big[w\big(\omega(\vec{x})|\rndVar{M}=1\big)y\Big(\vec{x}|\rndVar{M}=1 \Big)P(\rndVar{M}=1)\Big]}{ \sum_{{{m\in\mathbb{M}}}} \exp\Big[w\big(\omega(\vec{x})|\rndVar{M}=m\big)y\Big(\vec{x}|\rndVar{M}=m \Big)P(\rndVar{M}=m)\Big]} -0.5.
\end{align}

\subsection{Probability Estimation}\label{sec:Proopsed:ProbEstim}

In the previous section, a set of potential functions has been presented. For readability purposes, all necessary probability distributions were assumed to be known. Unfortunately, in real-world classification problems, these distributions remain unknown, and they have to be estimated using the training data. This section describes the techniques used to estimate the probabilities needed by the potential function. 

Prior class probabilities $P(\rndVar{M}=m)$ are estimated using the following formula: 
\begin{align}\label{eq:PriorProbEstim}
 \hat{P}(\rndVar{M}=m) &= \frac{|\mathcal{T}^{(m)}|}{|\mathcal{T}|},
\end{align}
where $\mathcal{T}^{(m)}$ is a subset of the training set containing objects which belong to class $m$:
\begin{equation}\label{eq:trainSetSplit} 
\mathcal{T}^{(m)}=\left\{({\vec{x}}^{(k)},m^{(k)})| m^{(k)} = m\right\}.
\end{equation}

To calculate the potential function $\newPotential(\vec{x})$, we need to estimate the probability distribution $y(\vec{x})$. Random variable $\rndVar{Y}$ has also been assumed to follow the multivariate normal distribution. Consequently, we use the maximum likelihood estimator to find the parameters of the distribution~\cite{Pan2002}.

For the conditional distribution $y(\vec{x}|\rndVar{M}=m)$, we considered two estimation procedures to be compared during the experimental study:
\begin{itemize}
 \item We assumed that the underlying random variable follows the multivariate Gaussian distribution. To estimate the conditional probability density function, we used the maximum likelihood estimator~\cite{Pan2002}.
 
 \item To make no assumptions about the shape of the distribution, we employed a nonparametric kernel estimator. To avoid using multidimensional kernels, we used the Naive Bayes assumption about the variables~\cite{Kulczycki2008,Wglarczyk2018}. In our work, we also decided to select the bandwidth using Silverman's rule of thumb~\cite{silverman1986density}
\end{itemize}

\subsection{Toy Examples}\label{sec:Proopsed:ToyExamples}

\begin{figure}[tb]
 \centering
  \includegraphics[width = 0.8\textwidth, height=.3\textheight, keepaspectratio]{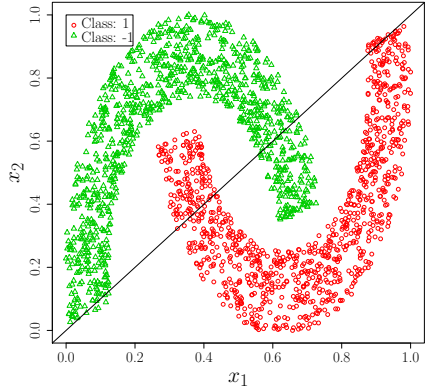}
  \caption{The linear decision boundary created by the Nearest Centroid classifier for binary, two-dimensional, banana-shaped data. Objects belonging to the first class have been marked using red circles, whereas points belonging to the other class have been marked using green triangles. The plot also shows the decision plane generated by the Nearest Centroid classifier.  \label{fig:LinClassSim2}}
\end{figure}

\begin{figure}[tb]
 \centering
  \includegraphics[width = 0.8\textwidth, height=.3\textheight, keepaspectratio]{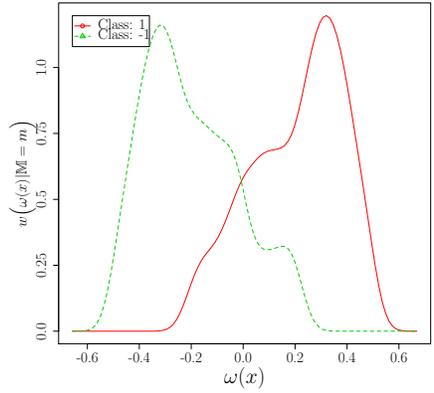}
  \caption{Class conditioned probability density functions estimated using the kernel estimator  \label{fig:KernProb}}
\end{figure}

In this section, the process of potential function building is visualized using a simple two-dimensional data set shown in \figurename~\ref{fig:LinClassSim2}. The decision boundary, shown in \figurename~\ref{fig:LinClassSim2},  is generated using the Nearest Centroid classifier~\cite{Kuncheva1998}. After obtaining the decision boundary, the conditional probability density functions $w\big(\omega(\vec{x})|\rndVar{M}=m\big)$ are estimated. The result is shown in \figurename~\ref{fig:KernProb}.

\begin{figure}[tb]
 \centering
  \includegraphics[width = 0.8\textwidth, height=.3\textheight, keepaspectratio]{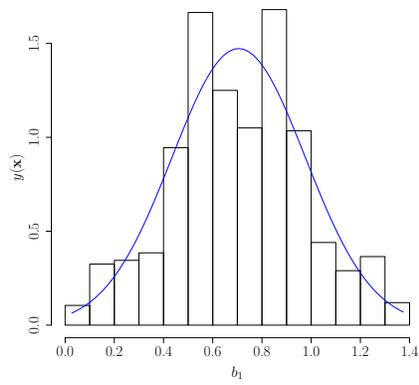}
  \caption{The probability density function $f(\vec{X})$ estimated using the maximum likelihood estimator. The histogram of values is also shown. \label{fig:p1BaseProb}}
\end{figure}

Now, the process of calculating the potential function $\newPotential(\vec{x})$ is shown.  First, the probability density function $y(\vec{x})$ is estimated. The result is shown in \figurename~\ref{fig:p1BaseProb}. Then the potential function $\newPotential(\vec{x})$ is calculated. It is visualised \figurename~\ref{fig:p1Pot}.

\begin{figure}[tb]
 \centering
  \includegraphics[width = 0.8\textwidth, height=.3\textheight, keepaspectratio]{\ptFiguresDirectory{pot1Potential}}
  \caption{The visualisation of the potential function $\newPotential(\vec{X})$.\label{fig:p1Pot}}
\end{figure}

\begin{figure}[tb]
 \centering
  \includegraphics[width = 0.8\textwidth, height=.3\textheight, keepaspectratio]{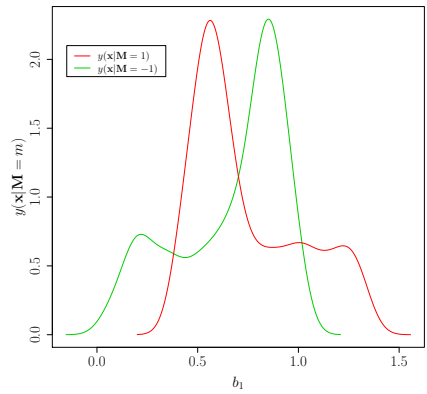}
  \caption{Probability density functions $f(\vec{X}|\rndVar{M}=m)$ estimated using the kernel estimator. \label{fig:p2BaseProbs}}
\end{figure}

Now, the process of calculating the potential function $\newPotential_{2}(\vec{x})$ is shown. First the conditional probability density functions $f(\vec{X}|\rndVar{M}=m)$ are estimated. The reault is shown in \figurename~\ref{fig:p2BaseProbs}. The potential function $\newPotential_2(\vec{x})$ is then calculated according to~\eqref{eq:Potential2}. It is visualised in~\figurename~\ref{fig:p2Pot}.

\begin{figure}[tb]
 \centering
  \includegraphics[width = 0.8\textwidth, height=.3\textheight, keepaspectratio]{\ptFiguresDirectory{pot2Potential}}
  \caption{The visualisation of the potential function $\newPotential_{2}(\vec{X})$.\label{fig:p2Pot}}
\end{figure}

\FloatBarrier
\section{Experimental Evaluation}\label{sec:ExpEval}

The main goal of the experimental evaluation is to answer the following research questions: 
\begin{itemize}
 \item[RQ1:]Does the utilization of the information about the point distribution along the basis of the decision plane significantly impact the classification quality achieved by the ensemble? 
 \item[RQ2:]Do the new formulated potential functions allow to improve the classification quality achieved by the ensemble? 
 \item [RQ3:] How is the ensemble utilizing the newly proposed potential functions doing compared to the ensemble created using Naive Bayes classifier which uses the class conditional probability calculated in a similar way. 
\end{itemize}

\subsection{Setup}\label{sec:ExpEval:setup}

\tablename~\ref{tab:BenchmarkSetsCharacteristics} displays the collection of the $70$ benchmark sets that were used during the experimental evaluation of the proposed methods. The table is divided into two sections.  Each section is organized as follows. The first column contains the names of the datasets. The remaining ones contain the set-specific characteristics of the benchmark sets: the number of instances in the dataset $|S|$; dimensionality of the input space $d$; the number of classes $C$ and the average imbalance ratio $\mathrm{IR}$, respectively.
{
\setlength\tabcolsep{1.9pt}%
\def\arraystretch{0.8}%
\begin{table}
 \centering\scriptsize
 \caption{The characteristics of the benchmark sets}\label{tab:BenchmarkSetsCharacteristics}
 \begin{tabular}{l cccc@{\hskip 0.2in}  l cccc}
Name&$|S|$&$d$&$C$&$\mathrm{IR}$&Name&$|S|$&$d$&$C$&$\mathrm{IR}$\\
\cmidrule(rr){1-5}  \cmidrule(lr){6-10}
abalone&4174&10&28&162.59&mammographic&830&5&2&1.03\\
adult&45222&103&2&2.02&marketing&6876&13&9&1.80\\
appendicitis&106&7&2&2.52&monk-2&432&6&2&1.06\\
australian&690&18&2&1.12&movement\_libras&360&90&15&1.00\\
automobile&159&61&6&4.30&mushroom&5644&92&2&1.31\\
balance&625&4&3&2.63&newthyroid&215&5&3&3.43\\
banana&5300&2&2&1.12&nursery&12960&26&5&435.25\\
bands&365&19&2&1.35&optdigits&5620&64&10&1.02\\
breast&277&38&2&1.71&page-blocks&5472&10&5&58.12\\
bupa&345&6&2&1.19&penbased&10992&16&10&1.04\\
car&1728&21&4&10.08&phoneme&5404&5&2&1.70\\
chess&3196&38&2&1.05&pima&768&8&2&1.43\\
cleveland&297&13&5&5.08&post-operative&87&21&3&21.86\\
coil2000&9822&85&2&8.38&ring&7400&20&2&1.01\\
connect-4&67557&126&3&3.52&saheart&462&9&2&1.44\\
contraceptive&1473&9&3&1.37&satimage&6435&36&6&1.66\\
crx&653&42&2&1.10&segment&2310&19&7&1.00\\
dermatology&358&34&6&2.43&shuttle&57999&9&7&1326.03\\
ecoli&336&7&8&23.56&sonar&208&60&2&1.07\\
fars&100968&362&8&610.12&spambase&4597&57&2&1.27\\
flare&1066&37&6&2.90&spectfheart&267&44&2&2.43\\
german&1000&59&2&1.67&splice&3190&287&3&1.77\\
glass&214&9&6&3.91&tae&151&5&3&1.03\\
haberman&306&3&2&1.89&texture&5500&40&11&1.00\\
hayes-roth&160&4&3&1.37&thyroid&7200&21&3&19.76\\
heart&270&13&2&1.13&tic-tac-toe&958&27&2&1.44\\
hepatitis&80&19&2&3.08&titanic&2201&3&2&1.55\\
housevotes&232&16&2&1.07&twonorm&7400&20&2&1.00\\
ionosphere&351&33&2&1.39&vehicle&846&18&4&1.03\\
iris&150&4&3&1.00&vowel&990&13&11&1.00\\
kr-vs-k&28056&40&18&20.96&wdbc&569&30&2&1.34\\
led7digit&500&7&10&1.16&wine&178&13&3&1.23\\
letter&20000&16&26&1.06&wisconsin&683&9&2&1.43\\
lymphography&148&38&4&15.77&yeast&1484&8&10&17.08\\
magic&19020&10&2&1.42&zoo&101&21&7&4.84\\

\end{tabular}
\end{table}
}

The datasets were taken from the Keel~\footnote{\url{https://sci2s.ugr.es/keel/category.php?cat=clas}} repository.
The datasets are also available in our repository~\footnote{Removed due to the double-blind review.}

During the dataset preprocessing stage, a few transformations on the datasets were applied. The PCA method~\cite{topolski2020modified} was applied and the percentage of covered variance was set to $0.95$. The attributes were also normalized to have zero mean and unit variance.

In the experimental study we conducted, the proposed potential functions were used to combine the predictions produced by a homogeneous ensemble of classifiers. The homogeneous ensembles were created using a bagging approach~\cite{Skurichina1998}. The generated ensembles consist of 11 classifiers learned by using the bagging method. Each bagging sample contains $80\%$ of the number of instances from the original dataset. 

For each of the kernel estimators used, the kernel bandwidth was selected using the Silverman rule~\cite {silverman1986density}. The Gaussian kernel is used.

During the experiment, the following ensembles were considered:
\begin{itemize}
 \item $\psi_{\mathrm{NB}}$ -- The ensemble created using Naive Bayes classifier~\cite{Hand2001}. 
 
 
 \item $\psi_{\mathrm{KE}}$ -- The ensemble in which base classifiers are combined according to approach proposed in~\cite{Trajdos2020a}. See also equation~\eqref{eq:pot1}. 
 \item $\psi_{\mathrm{KA}}$ -- The ensemble in which base classifiers are combined using the potential function defined in~\eqref{eq:newPotential1}.
 \item $\psi_{\mathrm{KB}}$ -- The ensemble in which base classifiers are combined using the potential function defined in~\eqref{eq:potN2}. The parametric gaussian estimator is used.
 \item $\psi_{\mathrm{KC}}$ -- The ensemble in which base classifiers are combined using the potential function defined in~\eqref{eq:potN2}. The kernel estimator is used.
\end{itemize}

The following base classifiers were used to build the above-mentioned ensembles (Except for Naive Bayes ensemble):
\begin{itemize}
 \item $\psi_{\mathrm{FLDA}}$ -- Fisher~LDA~\cite{McLachlan1992},
 \item $\psi_{\mathrm{LR}}$ -- Logistic regression classifier~\cite{Devroye1996},  
 \item $\psi_{\mathrm{MLP}}$ -- single layer MLP classifier~\cite{gurney1997an},
 \item $\psi_{\mathrm{NC}}$ -- nearest centroid (Nearest Prototype)~\cite{Kuncheva1998} with the class-specific Euclidean distance,
 \item $\psi_{\mathrm{SVM}}$ -- SVM classifier with linear kernel (no kernel)~\cite{Cortes1995}.
\end{itemize}

The classifiers used were implemented in the WEKA framework~\cite{Hall2009}. If not stated otherwise, the classifier parameters were set to their defaults. The multiclass problems were dealt with using One-vs-One decomposition~\cite{Hllermeier2010}. The source code of the proposed algorithms is available online~\footnote{Removed due to the double-blind review.}

To evaluate the proposed methods, six classification quality criteria are used: 
\begin{itemize}
 \item Macro-averaged:
 \begin{itemize}
  \item  false discovery rate  ($1- \mathrm{precision}$, $\fdr$);
  \item false negative rate ($1-\mathrm{recall}$, $\fnr$);
  \item Matthews correlation coefficient($\mcc$)
 \end{itemize}
 \item Micro-averaged:
 \begin{itemize}
  \item false discovery rate  ($1- \mathrm{precision}$, $\fdr$);
  \item false negative rate ($1-\mathrm{recall}$, $\fnr$);
 \item  Matthews correlation coefficient($\mcc$)
 \end{itemize}

\end{itemize}
Macro and micro-averaged measures were used to assess the performance for the majority and minority classes. This is because the macro-averaged measures are more sensitive to the performance for minority classes~\cite{Sokolova2009}. The criteria are bounded in the interval $[0,1]$, where zero denotes the best classification quality. To maintain consistency, the results obtained using the $\mcc$ criterion are also transformed to fit the above-mentioned properties.

The experimental procedure was conducted using the ten-fold cross-validation procedure. The data folds were generated using methods implemented in WEKA software. The random seed used to generate them is zero.

Following the recommendation of~\cite{garcia2008extension} the statistical significance of the obtained results was assessed using the two-step procedure. The first step was to perform the Iman-Davenport test~\cite{garcia2008extension} for each quality criterion separately. Since multiple criteria were employed, the family-wise errors (\FWER{}) should be controlled~\cite{Bergmann1988}. To do so, the Bergmann-Hommel~\cite{Bergmann1988} procedure of controlling \FWER{} of the conducted Iman-Davenport tests was employed. When the Iman-Davenport test shows that there is a significant difference within the group of classifiers, the Bergmann-Hommel post hoc test is applied~\cite{garcia2008extension,Bergmann1988}. For all tests, the significance level was set to $\alpha=0.05$.

\section{Results and Discussion}\label{sec:ResAndDisc}

\begin{figure}
\centering
\includegraphics[width=.7\textwidth, height=.3\textheight, keepaspectratio]{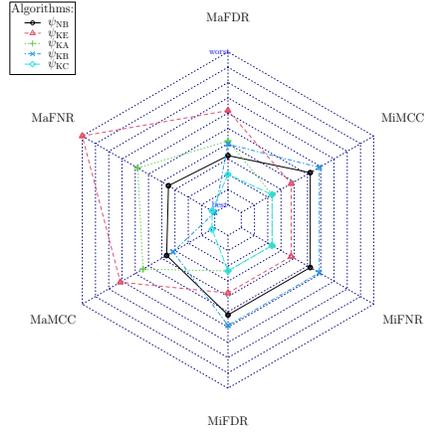}
 \caption{The radar plot for the ensembles based on $\psi_{\mathrm{FLDA}}$ }
 \label{fig:radarPlot:FLDA}
\end{figure}

\begin{figure}
\centering
\includegraphics[width=.7\textwidth, height=.3\textheight, keepaspectratio]{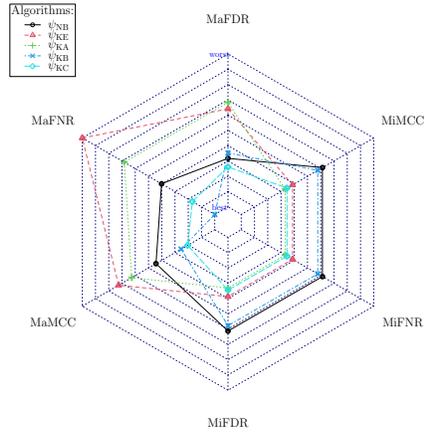}
 \caption{The radar plot for the ensembles based on $\psi_{\mathrm{LR}}$ }
 \label{fig:radarPlot:LR}
\end{figure}

\begin{figure}
\centering
\includegraphics[width=.7\textwidth, height=.3\textheight, keepaspectratio]{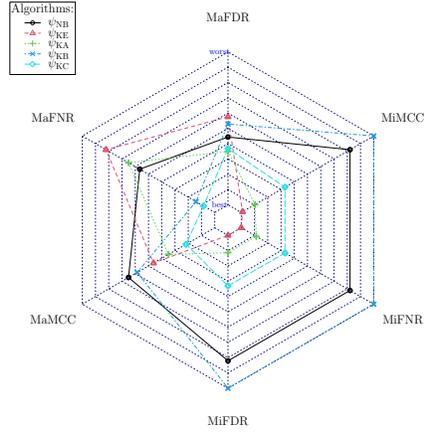}
 \caption{The radar plot for the ensembles based on $\psi_{\mathrm{MLP}}$ }
 \label{fig:radarPlot:MLP}
\end{figure}

\begin{figure}
\centering
\includegraphics[width=.7\textwidth, height=.3\textheight, keepaspectratio]{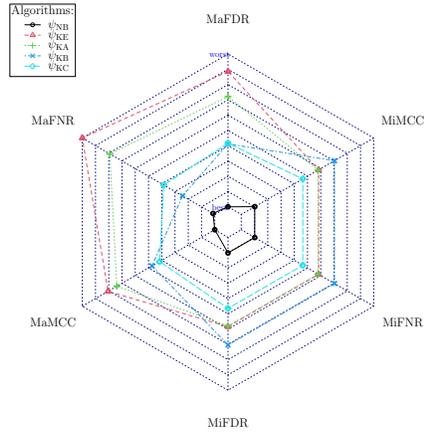}
 \caption{The radar plot for the ensembles based on $\psi_{\mathrm{NC}}$ }
 \label{fig:radarPlot:NC}
\end{figure}

\begin{figure}
\centering
\includegraphics[width=.7\textwidth, height=.3\textheight, keepaspectratio]{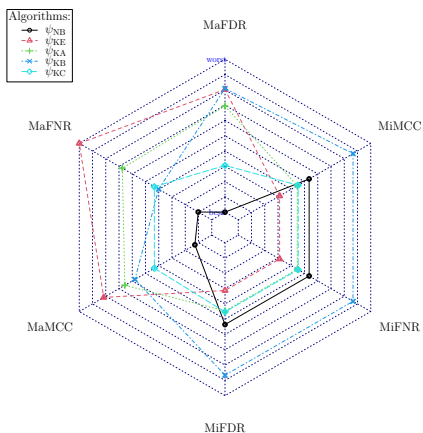}
 \caption{The radar plot for the ensembles based on $\psi_{\mathrm{SVM}}$ }
 \label{fig:radarPlot:SVM}
\end{figure}

To compare multiple algorithms on multiple benchmark sets, the average rank approach is used. In this approach, the winning algorithm achieves a rank equal to '1', the second achieves a rank equal to '2', and so on. In the case of ties, the ranks of algorithms that achieve the same results are averaged. To provide a visualization of the average ranks, radar plots are employed. In the radar plot, each of the radially arranged axes represents one quality criterion. In the plots, the data is visualized in such a way that the lowest ranks are closer to the centre of the graph. Consequently, higher ranks are placed near the outer ring of the graph. Graphs are also scaled so that the inner ring represents the lowest rank recorded for the analyzed set of classifiers, and the outer ring is equal to the highest recorded rank. The radar plots are presented in Figures~\ref{fig:radarPlot:FLDA} -- \ref{fig:radarPlot:SVM}.

The numerical results are given in \tablename~\ref{table:ensembleFLDA}~to~ \ref{table:ensembleSVM}. Each table is structured as follows. The first row contains the names of the investigated algorithms. Then, the table is divided into six sections -- one section is related to a single evaluation criterion. The first row of each section is the name of the quality criterion investigated in the section. The second row shows the p-value of the Iman-Davenport test. The third one shows the average ranks achieved by algorithms. The following rows show p-values resulting from the post hoc test. The p-value equal to $.000$ informs that the p-values are lower than $10^{-3}$ and p-value equal to $1.00$ informs that the value is higher than $0.999$. P-values lower (or equal) than $\alpha$ are bolded. Consequently, the bolded results show that there is a significant difference between classifiers.

Let us begin with the analysis of differences between $\psi_{\mathrm{KE}}$ and its modifications that allow us to incorporate the information about points spread along the decision plane basis ($\psi_{\mathrm{KA}}$, $\psi_{\mathrm{KB}}$, and $\psi_{\mathrm{KC}}$ respectively). The conducted statistical analysis reports only a few significant differences between these methods.  

Most of the differences are observed for the macro-averaged $\fnr$ criterion. In terms of this criterion $\psi_{\mathrm{KB}}$ and $\psi_{\mathrm{KC}}$ tend to be better than $\psi_{\mathrm{KE}}$. For this criterion, the average ranks achieved by the proposed methods tend to be lower than the ranks achieved by $\psi_{\mathrm{KE}}$. For the macro-averaged $\fdr$ measure, almost no significant differences have been reported. For this criterion, only one significant difference is reported for $\psi_{\mathrm{NC}}$ base classifier.  What is more, the order (according to the average ranks) of classifiers depends on the base classifier used. It means that for the minority classes, the modified methods ($\psi_{\mathrm{KB}}$ and $\psi_{\mathrm{KC}}$) improve the recall without harming the precision. Unfortunately, the overall classification quality expressed in terms of the macro-averaged $\mcc$ criterion has not been significantly improved. However, we may observe that the averaged ranks for the proposed methods tend to be lower for this criterion. This is especially for $\psi_{\mathrm{KB}}$ and $\psi_{\mathrm{KC}}$.  This trend is observed for all base classifiers. It means that including the information about the point spread along the hyperplane basis allows obtaining some improvement. Utilizing the class-specific densities $y(\vec{x}|\rndVar{M}=m)$ causes higher differences in average ranks. It means that using class-specific densities gives better results than using a global density $P(\rndVar{X}=x)$.

For the micro-averaged criteria, significant differences are observed only for a sole base classifier. The difference is observed for the $\psi_{\mathrm{MLP}}$ base classifier and all micro-averaged criteria. However, in this case, the result is not so strong because the p-value resulted from the Iman Davenport test is above the significance level $\alpha$. Apart from that, the results of the post-hoc test show that the $\psi_{\mathrm{KE}}$ classifier is significantly better than $\psi_{\mathrm{KB}}$ classifier. What is more, the $\psi_{\mathrm{KB}}$ classifier tends to achieve higher ranks than $\psi_{\mathrm{KE}}$ for all micro averaged criteria and base classifiers. It means that $\psi_{\mathrm{KB}}$ classifier may be weaker when classifying the examples from the majority classes. For $\psi_{\mathrm{KA}}$ and $\psi_{\mathrm{KC}}$, on the other hand, the average ranks for micro-averaged criteria are lower than the ranks calculated for $\psi_{\mathrm{KE}}$. This result shows that for the proposed method, the choice of the probability estimation method is fairly important. The nonparametric estimator seems to be a better choice than a parametric one related to the arbitrary chosen distribution (The Gaussian one in this study). This is likely due to the ability of the kernel estimator to provide a better estimation of the multimodal probability density.

Finally, let us compare $\psi_{\mathrm{KA}}$, $\psi_{\mathrm{KB}}$ and $\psi_{\mathrm{KC}}$  classifies and the ensemble built using the Naive Bayes algorithm ($\psi_{\mathrm{NB}}$). This comparison needs to be made since the algorithms use a similar approach to estimating the multidimensional probability distribution as the Naive Bayes algorithm does.

First of all, for the macro-averaged $\fnr$ and $\fdr$ measures, the $\psi_{\mathrm{NB}}$ classifier significantly outperforms the remaining classifiers for three out of five base classifiers. What is more, the $\psi_{\mathrm{NB}}$ is also better in terms of the macro-averaged $\mcc$ classifier for the nearest centroid base classifier. This is also true for the $\psi_{\mathrm{KC}}$ classifier that uses almost the same procedure for estimation probability. The probable reasons for these differences are twofold. The first reason is that the $\psi_{\mathrm{KC}}$ classifier uses one-vs-one decomposition to deal with multiclass problems whereas $\psi_{\mathrm{NB}}$ can handle multiclass problems directly. The second reason is that $\psi_{\mathrm{NB}}$ estimates the probabilities using original attributes which are, due to applied PCA transformation, uncorrelated. $\psi_{\mathrm{KC}}$ on the other hand uses an input space spanned by the normal vector and decision hyperplane basis. The literature shows that applying $\psi_{\mathrm{NB}}$ on uncorrelated attributes gives better results~\cite{Fan2007}.

For the micro-averaged measures, on the other hand, no significant difference is observed. It means that, when dealing with the majority classes, the investigated classifiers offer comparable classification quality.

{
\setlength\tabcolsep{1.0pt}%
 \def\arraystretch{0.5}%
\begin{table}[htb]
\centering\tiny
\caption{Statistical evaluation: the post-hoc test for the ensembles based on the FLDA classifier.\label{table:ensembleFLDA}}
\begin{tabular}{c@{\hskip 0.05in} ccccc@{\hskip 0.05in} ccccc@{\hskip 0.05in}ccccc}
  & $\psi_{\mathrm{NB}}$ & $\psi_{\mathrm{KE}}$ &  $\psi_{\mathrm{KA}}$ & $\psi_{\mathrm{KB}}$ & $\psi_{\mathrm{KC}}$ & $\psi_{\mathrm{NB}}$ &  $\psi_{\mathrm{KE}}$ &  $\psi_{\mathrm{KA}}$ & $\psi_{\mathrm{KB}}$ & $\psi_{\mathrm{KC}}$ & $\psi_{\mathrm{NB}}$ &  $\psi_{\mathrm{KE}}$ &  $\psi_{\mathrm{KA}}$ & $\psi_{\mathrm{KB}}$ & $\psi_{\mathrm{KC}}$ \\ 
  \cmidrule(lr){ 2-6 } \cmidrule(lr){ 7-11 } \cmidrule(lr){ 12-16 }
Nam.&\multicolumn{5}{c}{MaFDR}&\multicolumn{5}{c}{MaFNR}&\multicolumn{5}{c}{MaMCC}\\
ImD.&\multicolumn{5}{c}{1.000e+00}&\multicolumn{5}{c}{\textbf{1.253e-03}}&\multicolumn{5}{c}{4.155e-01}\\
Rank & 2.929 & 3.236 & 3.029 & 3.007 & 2.800 & 2.957 & 3.636 & 3.200 & 2.593 & 2.614 & 2.971 & 3.336 & 3.157 & 2.921 & 2.614 \\ 
  $\psi_{\mathrm{NB}}$ &  & 1.00 & 1.00 & 1.00 & 1.00 &  & \textbf{.044} & .727 & .519 & .519 &  & .727 & 1.00 & 1.00 & .727 \\ 
  $\psi_{\mathrm{KE}}$ &  &  & 1.00 & 1.00 & 1.00 &  &  & .412 & \textbf{.001} & \textbf{.001} &  &  & 1.00 & .727 & .069 \\ 
  $\psi_{\mathrm{KA}}$ &  &  &  & 1.00 & 1.00 &  &  &  & .139 & .139 &  &  &  & 1.00 & .253 \\ 
  $\psi_{\mathrm{KB}}$ &  &  &  &  & 1.00 &  &  &  &  & .936 &  &  &  &  & .727 \\ 
   \cmidrule(lr){ 2-6 } \cmidrule(lr){ 7-11 } \cmidrule(lr){ 12-16 }
Nam.&\multicolumn{5}{c}{MiFDR}&\multicolumn{5}{c}{MiFNR}&\multicolumn{5}{c}{MiMCC}\\
ImD.&\multicolumn{5}{c}{1.000e+00}&\multicolumn{5}{c}{1.000e+00}&\multicolumn{5}{c}{1.000e+00}\\
Rank & 3.136 & 2.986 & 2.836 & 3.207 & 2.836 & 3.136 & 2.986 & 2.836 & 3.207 & 2.836 & 3.136 & 2.986 & 2.836 & 3.207 & 2.836 \\ 
  $\psi_{\mathrm{NB}}$ &  & 1.00 & 1.00 & 1.00 & 1.00 &  & 1.00 & 1.00 & 1.00 & 1.00 &  & 1.00 & 1.00 & 1.00 & 1.00 \\ 
  $\psi_{\mathrm{KE}}$ &  &  & 1.00 & 1.00 & 1.00 &  &  & 1.00 & 1.00 & 1.00 &  &  & 1.00 & 1.00 & 1.00 \\ 
  $\psi_{\mathrm{KA}}$ &  &  &  & 1.00 & 1.00 &  &  &  & 1.00 & 1.00 &  &  &  & 1.00 & 1.00 \\ 
  $\psi_{\mathrm{KB}}$ &  &  &  &  & 1.00 &  &  &  &  & 1.00 &  &  &  &  & 1.00 \\ 
  
  \end{tabular}
\end{table}
}

{
\setlength\tabcolsep{1.0pt}%
 \def\arraystretch{0.5}%
\begin{table}[htb]
\centering\tiny
\caption{Statistical evaluation: the post-hoc test for the ensembles based on the LR classifier.\label{table:ensembleLR}}
\begin{tabular}{c@{\hskip 0.05in} ccccc@{\hskip 0.05in} ccccc@{\hskip 0.05in}ccccc}
  & $\psi_{\mathrm{NB}}$ & $\psi_{\mathrm{KE}}$ &  $\psi_{\mathrm{KA}}$ & $\psi_{\mathrm{KB}}$ & $\psi_{\mathrm{KC}}$ & $\psi_{\mathrm{NB}}$ &  $\psi_{\mathrm{KE}}$ &  $\psi_{\mathrm{KA}}$ & $\psi_{\mathrm{KB}}$ & $\psi_{\mathrm{KC}}$ & $\psi_{\mathrm{NB}}$ &  $\psi_{\mathrm{KE}}$ &  $\psi_{\mathrm{KA}}$ & $\psi_{\mathrm{KB}}$ & $\psi_{\mathrm{KC}}$ \\ 
  \cmidrule(lr){ 2-6 } \cmidrule(lr){ 7-11 } \cmidrule(lr){ 12-16 }
Nam.&\multicolumn{5}{c}{MaFDR}&\multicolumn{5}{c}{MaFNR}&\multicolumn{5}{c}{MaMCC}\\
ImD.&\multicolumn{5}{c}{1.000e+00}&\multicolumn{5}{c}{\textbf{4.966e-03}}&\multicolumn{5}{c}{1.000e+00}\\
Rank & 2.864 & 3.193 & 3.236 & 2.900 & 2.807 & 2.950 & 3.557 & 3.236 & 2.543 & 2.714 & 2.993 & 3.279 & 3.179 & 2.800 & 2.750 \\ 
  $\psi_{\mathrm{NB}}$ &  & 1.00 & 1.00 & 1.00 & 1.00 &  & .092 & .570 & .511 & .570 &  & 1.00 & 1.00 & 1.00 & 1.00 \\ 
  $\psi_{\mathrm{KE}}$ &  &  & 1.00 & 1.00 & 1.00 &  &  & .511 & \textbf{.001} & \textbf{.010} &  &  & 1.00 & .480 & .480 \\ 
  $\psi_{\mathrm{KA}}$ &  &  &  & 1.00 & 1.00 &  &  &  & .057 & .153 &  &  &  & .653 & .653 \\ 
  $\psi_{\mathrm{KB}}$ &  &  &  &  & 1.00 &  &  &  &  & .570 &  &  &  &  & 1.00 \\ 
   \cmidrule(lr){ 2-6 } \cmidrule(lr){ 7-11 } \cmidrule(lr){ 12-16 }
Nam.&\multicolumn{5}{c}{MiFDR}&\multicolumn{5}{c}{MiFNR}&\multicolumn{5}{c}{MiMCC}\\
ImD.&\multicolumn{5}{c}{1.000e+00}&\multicolumn{5}{c}{1.000e+00}&\multicolumn{5}{c}{1.000e+00}\\
Rank & 3.164 & 2.936 & 2.879 & 3.129 & 2.893 & 3.164 & 2.936 & 2.879 & 3.129 & 2.893 & 3.164 & 2.936 & 2.879 & 3.129 & 2.893 \\ 
  $\psi_{\mathrm{NB}}$ &  & 1.00 & 1.00 & 1.00 & 1.00 &  & 1.00 & 1.00 & 1.00 & 1.00 &  & 1.00 & 1.00 & 1.00 & 1.00 \\ 
  $\psi_{\mathrm{KE}}$ &  &  & 1.00 & 1.00 & 1.00 &  &  & 1.00 & 1.00 & 1.00 &  &  & 1.00 & 1.00 & 1.00 \\ 
  $\psi_{\mathrm{KA}}$ &  &  &  & 1.00 & 1.00 &  &  &  & 1.00 & 1.00 &  &  &  & 1.00 & 1.00 \\ 
  $\psi_{\mathrm{KB}}$ &  &  &  &  & 1.00 &  &  &  &  & 1.00 &  &  &  &  & 1.00 \\ 
  
  \end{tabular}
\end{table}
}

{
\setlength\tabcolsep{1.0pt}%
 \def\arraystretch{0.5}%
\begin{table}[htb]
\centering\tiny
\caption{Statistical evaluation: the post-hoc test for the ensembles based on the MLP classifier.\label{table:ensembleMLP}}
\begin{tabular}{c@{\hskip 0.05in} ccccc@{\hskip 0.05in} ccccc@{\hskip 0.05in}ccccc}
  & $\psi_{\mathrm{NB}}$ & $\psi_{\mathrm{KE}}$ &  $\psi_{\mathrm{KA}}$ & $\psi_{\mathrm{KB}}$ & $\psi_{\mathrm{KC}}$ & $\psi_{\mathrm{NB}}$ &  $\psi_{\mathrm{KE}}$ &  $\psi_{\mathrm{KA}}$ & $\psi_{\mathrm{KB}}$ & $\psi_{\mathrm{KC}}$ & $\psi_{\mathrm{NB}}$ &  $\psi_{\mathrm{KE}}$ &  $\psi_{\mathrm{KA}}$ & $\psi_{\mathrm{KB}}$ & $\psi_{\mathrm{KC}}$ \\ 
  \cmidrule(lr){ 2-6 } \cmidrule(lr){ 7-11 } \cmidrule(lr){ 12-16 }
Nam.&\multicolumn{5}{c}{MaFDR}&\multicolumn{5}{c}{MaFNR}&\multicolumn{5}{c}{MaMCC}\\
ImD.&\multicolumn{5}{c}{1.000e+00}&\multicolumn{5}{c}{4.497e-01}&\multicolumn{5}{c}{1.000e+00}\\
Rank & 2.993 & 3.093 & 2.921 & 3.057 & 2.936 & 3.086 & 3.279 & 3.150 & 2.764 & 2.721 & 3.150 & 3.007 & 2.921 & 3.100 & 2.821 \\ 
  $\psi_{\mathrm{NB}}$ &  & 1.00 & 1.00 & 1.00 & 1.00 &  & 1.00 & 1.00 & .691 & .691 &  & 1.00 & 1.00 & 1.00 & 1.00 \\ 
  $\psi_{\mathrm{KE}}$ &  &  & 1.00 & 1.00 & 1.00 &  &  & 1.00 & .371 & .371 &  &  & 1.00 & 1.00 & 1.00 \\ 
  $\psi_{\mathrm{KA}}$ &  &  &  & 1.00 & 1.00 &  &  &  & .653 & .653 &  &  &  & 1.00 & 1.00 \\ 
  $\psi_{\mathrm{KB}}$ &  &  &  &  & 1.00 &  &  &  &  & 1.00 &  &  &  &  & 1.00 \\ 
   \cmidrule(lr){ 2-6 } \cmidrule(lr){ 7-11 } \cmidrule(lr){ 12-16 }
Nam.&\multicolumn{5}{c}{MiFDR}&\multicolumn{5}{c}{MiFNR}&\multicolumn{5}{c}{MiMCC}\\
ImD.&\multicolumn{5}{c}{8.367e-02}&\multicolumn{5}{c}{8.367e-02}&\multicolumn{5}{c}{8.367e-02}\\
Rank & 3.279 & 2.657 & 2.743 & 3.414 & 2.907 & 3.279 & 2.657 & 2.743 & 3.414 & 2.907 & 3.279 & 2.664 & 2.736 & 3.414 & 2.907 \\ 
  $\psi_{\mathrm{NB}}$ &  & .120 & .135 & 1.00 & .329 &  & .120 & .135 & 1.00 & .329 &  & .129 & .129 & 1.00 & .329 \\ 
  $\psi_{\mathrm{KE}}$ &  &  & 1.00 & \textbf{.046} & 1.00 &  &  & 1.00 & \textbf{.046} & 1.00 &  &  & 1.00 & \textbf{.050} & 1.00 \\ 
  $\psi_{\mathrm{KA}}$ &  &  &  & .072 & 1.00 &  &  &  & .072 & 1.00 &  &  &  & .067 & 1.00 \\ 
  $\psi_{\mathrm{KB}}$ &  &  &  &  & .231 &  &  &  &  & .231 &  &  &  &  & .231 \\ 
  
  \end{tabular}
\end{table}
}

{
\setlength\tabcolsep{1.0pt}%
 \def\arraystretch{0.5}%
\begin{table}[htb]
\centering\tiny
\caption{Statistical evaluation: the post-hoc test for the ensembles based on the NC classifier.\label{table:ensembleNC}}
\begin{tabular}{c@{\hskip 0.05in} ccccc@{\hskip 0.05in} ccccc@{\hskip 0.05in}ccccc}
  & $\psi_{\mathrm{NB}}$ & $\psi_{\mathrm{KE}}$ &  $\psi_{\mathrm{KA}}$ & $\psi_{\mathrm{KB}}$ & $\psi_{\mathrm{KC}}$ & $\psi_{\mathrm{NB}}$ &  $\psi_{\mathrm{KE}}$ &  $\psi_{\mathrm{KA}}$ & $\psi_{\mathrm{KB}}$ & $\psi_{\mathrm{KC}}$ & $\psi_{\mathrm{NB}}$ &  $\psi_{\mathrm{KE}}$ &  $\psi_{\mathrm{KA}}$ & $\psi_{\mathrm{KB}}$ & $\psi_{\mathrm{KC}}$ \\ 
  \cmidrule(lr){ 2-6 } \cmidrule(lr){ 7-11 } \cmidrule(lr){ 12-16 }
Nam.&\multicolumn{5}{c}{MaFDR}&\multicolumn{5}{c}{MaFNR}&\multicolumn{5}{c}{MaMCC}\\
ImD.&\multicolumn{5}{c}{\textbf{1.866e-06}}&\multicolumn{5}{c}{\textbf{1.084e-09}}&\multicolumn{5}{c}{\textbf{1.184e-05}}\\
Rank & 2.214 & 3.643 & 3.379 & 2.879 & 2.886 & 2.236 & 3.829 & 3.493 & 2.607 & 2.836 & 2.214 & 3.514 & 3.407 & 2.979 & 2.886 \\ 
  $\psi_{\mathrm{NB}}$ &  & \textbf{.000} & \textbf{.000} & \textbf{.048} & \textbf{.048} &  & \textbf{.000} & \textbf{.000} & .329 & .099 &  & \textbf{.000} & \textbf{.000} & \textbf{.017} & \textbf{.048} \\ 
  $\psi_{\mathrm{KE}}$ &  &  & .645 & \textbf{.025} & \textbf{.025} &  &  & .418 & \textbf{.000} & \textbf{.001} &  &  & 1.00 & .135 & .112 \\ 
  $\psi_{\mathrm{KA}}$ &  &  &  & .184 & .184 &  &  &  & \textbf{.003} & \textbf{.028} &  &  &  & .153 & .153 \\ 
  $\psi_{\mathrm{KB}}$ &  &  &  &  & .979 &  &  &  &  & .418 &  &  &  &  & 1.00 \\ 
   \cmidrule(lr){ 2-6 } \cmidrule(lr){ 7-11 } \cmidrule(lr){ 12-16 }
Nam.&\multicolumn{5}{c}{MiFDR}&\multicolumn{5}{c}{MiFNR}&\multicolumn{5}{c}{MiMCC}\\
ImD.&\multicolumn{5}{c}{\textbf{9.557e-03}}&\multicolumn{5}{c}{\textbf{9.557e-03}}&\multicolumn{5}{c}{\textbf{9.557e-03}}\\
Rank & 2.379 & 3.157 & 3.150 & 3.350 & 2.964 & 2.379 & 3.157 & 3.150 & 3.350 & 2.964 & 2.379 & 3.157 & 3.150 & 3.350 & 2.964 \\ 
  $\psi_{\mathrm{NB}}$ &  & \textbf{.021} & \textbf{.021} & \textbf{.003} & .114 &  & \textbf{.021} & \textbf{.021} & \textbf{.003} & .114 &  & \textbf{.021} & \textbf{.021} & \textbf{.003} & .114 \\ 
  $\psi_{\mathrm{KE}}$ &  &  & 1.00 & 1.00 & 1.00 &  &  & 1.00 & 1.00 & 1.00 &  &  & 1.00 & 1.00 & 1.00 \\ 
  $\psi_{\mathrm{KA}}$ &  &  &  & 1.00 & 1.00 &  &  &  & 1.00 & 1.00 &  &  &  & 1.00 & 1.00 \\ 
  $\psi_{\mathrm{KB}}$ &  &  &  &  & .894 &  &  &  &  & .894 &  &  &  &  & .894 \\ 
  
  \end{tabular}
\end{table}
}

{
\setlength\tabcolsep{1.0pt}%
 \def\arraystretch{0.5}%
\begin{table}[htb]
\centering\tiny
\caption{Statistical evaluation: the post-hoc test for the ensembles based on the SVM classifier.\label{table:ensembleSVM}}
\begin{tabular}{c@{\hskip 0.05in} ccccc@{\hskip 0.05in} ccccc@{\hskip 0.05in}ccccc}
  & $\psi_{\mathrm{NB}}$ & $\psi_{\mathrm{KE}}$ &  $\psi_{\mathrm{KA}}$ & $\psi_{\mathrm{KB}}$ & $\psi_{\mathrm{KC}}$ & $\psi_{\mathrm{NB}}$ &  $\psi_{\mathrm{KE}}$ &  $\psi_{\mathrm{KA}}$ & $\psi_{\mathrm{KB}}$ & $\psi_{\mathrm{KC}}$ & $\psi_{\mathrm{NB}}$ &  $\psi_{\mathrm{KE}}$ &  $\psi_{\mathrm{KA}}$ & $\psi_{\mathrm{KB}}$ & $\psi_{\mathrm{KC}}$ \\ 
  \cmidrule(lr){ 2-6 } \cmidrule(lr){ 7-11 } \cmidrule(lr){ 12-16 }
Nam.&\multicolumn{5}{c}{MaFDR}&\multicolumn{5}{c}{MaFNR}&\multicolumn{5}{c}{MaMCC}\\
ImD.&\multicolumn{5}{c}{\textbf{6.298e-03}}&\multicolumn{5}{c}{\textbf{1.400e-02}}&\multicolumn{5}{c}{1.908e-01}\\
Rank & 2.414 & 3.314 & 3.193 & 3.321 & 2.757 & 2.529 & 3.536 & 3.171 & 2.864 & 2.900 & 2.557 & 3.329 & 3.150 & 3.064 & 2.900 \\ 
  $\psi_{\mathrm{NB}}$ &  & \textbf{.007} & \textbf{.014} & \textbf{.007} & .798 &  & \textbf{.002} & .097 & .658 & .658 &  & \textbf{.039} & .159 & .231 & .798 \\ 
  $\psi_{\mathrm{KE}}$ &  &  & 1.00 & 1.00 & .208 &  &  & .658 & .072 & .097 &  &  & 1.00 & .968 & .653 \\ 
  $\psi_{\mathrm{KA}}$ &  &  &  & 1.00 & .208 &  &  &  & .751 & .751 &  &  &  & 1.00 & 1.00 \\ 
  $\psi_{\mathrm{KB}}$ &  &  &  &  & .208 &  &  &  &  & .894 &  &  &  &  & 1.00 \\ 
   \cmidrule(lr){ 2-6 } \cmidrule(lr){ 7-11 } \cmidrule(lr){ 12-16 }
Nam.&\multicolumn{5}{c}{MiFDR}&\multicolumn{5}{c}{MiFNR}&\multicolumn{5}{c}{MiMCC}\\
ImD.&\multicolumn{5}{c}{5.739e-01}&\multicolumn{5}{c}{5.739e-01}&\multicolumn{5}{c}{5.739e-01}\\
Rank & 3.014 & 2.764 & 2.914 & 3.386 & 2.921 & 3.014 & 2.764 & 2.914 & 3.386 & 2.921 & 3.014 & 2.764 & 2.914 & 3.386 & 2.921 \\ 
  $\psi_{\mathrm{NB}}$ &  & 1.00 & 1.00 & .658 & 1.00 &  & 1.00 & 1.00 & .658 & 1.00 &  & 1.00 & 1.00 & .658 & 1.00 \\ 
  $\psi_{\mathrm{KE}}$ &  &  & 1.00 & .201 & 1.00 &  &  & 1.00 & .201 & 1.00 &  &  & 1.00 & .201 & 1.00 \\ 
  $\psi_{\mathrm{KA}}$ &  &  &  & .466 & 1.00 &  &  &  & .466 & 1.00 &  &  &  & .466 & 1.00 \\ 
  $\psi_{\mathrm{KB}}$ &  &  &  &  & .466 &  &  &  &  & .466 &  &  &  &  & .466 \\ 
  
  \end{tabular}
\end{table}
}


\section{Conclusions}\label{sec:Conclusions}

In this paper, we proposed a few modifications of the algorithm proposed in~\cite{Trajdos2020a}. The modifications allow the aforementioned algorithm to utilize the information about the point spread along the decision plane basis. This information should allow the created ensemble to better view the competence regions of the employed base classifiers. Identifying competence regions should allow the ensemble to achieve better classification quality than the ensemble that does not use this information. 

We conducted a set of experiments using different base classifiers and a set of different quality measures to answer the formulated research questions. The experiments were conducted using 70 publicly available benchmark sets.

The experimental study carried out allowed us to provide the following answers to the research questions raised. 
\begin{itemize}
 \item[RQ1:] The utilization of the information about the point distribution along the basis of the decision plane has some impact on the classification quality obtained by the ensemble. 
 \item[RQ2:] The utilization of the new formulated potential functions improves the ensemble's classification quality only for the macro-averaged $\fnr$ criterion.  
 \item [RQ3:] The ensemble using the proposed new potential functions is comparable to the ensemble constructed using the Naive Bayes classifier in terms of four out of six criteria. For the remaining criteria, they are worse.  
\end{itemize}

The proposed modifications do not significantly outperform the initial approach. Consequently, our future research should be aimed at other techniques of improving the ensembles of linear classifiers. For example, a different ensemble building technique may be proposed.  


\FloatBarrier
\bibliography{bibliography}

\end{document}